# Fast Pedestrian Detection based on T-CENTRIST in infrared image

Hongyin Ni[1], Fengping Li[2]

(1. Northeast Electric Power University, School of Computer, No.169, Changchun Road, Jilin, P. R. China, 132012

2. College of Humanities Information Changchun University of Technology, No. 1016, FuZhi Road, Jilin, P. R. China,130122.)

**Abstract:** Pedestrian detection is a research hotspot and a difficult issue in the computer vision such as the Intelligent Surveillance System (ISS), the Intelligent Transport System (ITS), robotics, and automotive safety. However, the human body's position, angle, and dress in a video scene are complicated and changeable, which have a great influence on the detection accuracy. In this paper, through the analysis on the pros and cons of Census Transform Histogram (CENTRIST), a novel feature is presented for human detection——Ternary CENTRIST (T-CENTRIST). The T-CENTRIST feature takes the relationship between each pixel and its neighborhood pixels into account. Meanwhile, it also considers the relevancy among these neighborhood pixels. Therefore, the proposed feature description method can reflect the silhouette of pedestrian more adequately and accurately than that of CENTRIST. Second, we propose a fast pedestrian detection framework based on T-CENTRIST in infrared image, which introduces the idea of extended blocks and the integral image. Finally, experimental results verify the effectiveness of the proposed pedestrian detection method.

## 1 Introduction

Pedestrian detection can be understood as a method of determining if a pedestrian is contained in an input image or video stream. If so, the position information of the pedestrian is given [1,2]. In intelligent video surveillance [3], intelligent robotics, automatic (auxiliary) driving and ITS [4,5,6], pedestrian detection has a wide range of applications. However, because human body is a non-rigid target, the position, angle, and dress in a video scene are complicated and changeable. These factors have a great influence on the detection accuracy and lead to existing large-scale changes. At the same time, the processing time is extended seriously. In current researches, pedestrian detection methods focus on the following two aspects: feature extraction and detector structure [1,4,6，23,24,25,26,27,28].

In pedestrian feature extraction, it has made many impressive achievements. P. Viola et al. proposed the Haar feature for pedestrian detection trained by an Adaboost training algorithm to realize a pedestrian detection[7,8]. The processing time of this algorithm could achieve approximately 0.25s in an image with 360 × 240 pixels. In 2005, Dalal et al. presented Histogram of Oriented Gradient (HOG) feature, which could describe the tendency of pedestrian gradient and improve the accuracy of pedestrian detection[9]. However, the disadvantage is that HOG feature vectors have a large number of dimensions, affecting the processing speed to some extent. Bo Wu et al. put forward the edgelet feature for achieving a part-based pedestrian detection method [10]. The edgelet can describe contour changes of the local pedestrian effectively. However, due to dramatic changes in body contours, this method can only detect the front or back of the body, thus decreasing the accuracy of the method. Through researching Local Binary Pattern (LBP), Y. Mu et al. proposed Semantic-LBP and Fourier-LBP, which have been successfully applied to human detection, and have achieved better detection results [11]. However, the detection performance decreased significantly in the low-resolution image. In 2011, the CENTRIST feature was proposed by Wu et al. [12]. It was originally used for the scene classification. Subsequently, it was successfully applied to extract the features of human body [13]. With the help of integral image technology, the processing time of pedestrian detection is reduced greatly. In addition, for improving the accuracy of pedestrian detection, many experts and scholars used more information (like color, texture, edge and gradient) to detect human [14,15,16]. Although these methods can improve accuracy rate, the computational capacity involved is enormous.

After the pedestrian feature extraction, the next work is to construct a detector by using these features. Currently, there are two most commonly used machine-learning algorithms: Adaptive Boosting (Adaboost) and Support Vector Machines (SVMs) [17,18]. Adaboost is a greedy algorithm, which is known to obtain a simple classifier (i.e. weak classifier) by changing the distribution of the training sample set, and then to form the final classifier (i.e. strong classifier) after these weak classifiers are put together. However, if there are some difficult or rare samples in the training set, Adaboost is prone to over-adjust the problem, which will lead to a sharp decline in the performance of the strong classifier. SVM is a supervised learning model that analyzes data and recognizes patterns used for classification and regression analysis, whose core idea is to solve a quadratic programming problem. SVM can efficiently perform a non-linear classification by using what is called the kernel trick, and implicitly mapping their inputs into high-dimensional feature spaces. Therefore, it has a very wide range of applications in pattern recognition. In order to improve the detection accuracy, Y. Wang, X. Lin et al. proposed Deep Networks for Visual Recognition[29,30] and applied to person re-identification which achieved good results[31,32,33,34].

Although much progress has been made in pedestrian detection, the detection rate is still unable to meet the requirement of precision. Based on current pedestrian feature extraction algorithms, CENTRIST shows excellent detection performance. Therefore, in this paper, according to the pros and cons of CENTRIST, a novel feature description method——T-CENTRIST is proposed, which can be used to extract pedestrian features in an image and can better reflect the relationship among different adjacent pixels. Then a fast pedestrian detection framework by using T-CENTRIST is represented, which can improve pedestrian detection accuracy and reduce the false alarm rate significantly. At last, an evaluation of our approach on INRIA data set is made, and it shows significant improvements in contrast with some state-of-art methods.

The outline of the paper is as follows: Sec. 2 is the introduction of the Census Transform and CENTRIST. Next, a novel pedestrian feature called T-CENTRIST is proposed through the analysis on the pros and cons of CENTRIST. In Sec. 3, the overview of our pedestrian detection framework is introduced. The experimental results are shown in Sec. 4. The conclusion of this paper and future work are given in Sec. 5.

## 2 T-CENTRIST

### 2.1 Census Transform and CENTRIST

In 1996, Zabih et al. proposed Census Transform (CT), a non-parametric local relationship conversion method [20]. In the image processing, CT can be used to obtain the relationship between any given pixel and its neighboring pixels.

Suppose there is an image $I$, $\forall p \in I$, $I_p$ is the grayscale value of point $p$, and $\delta(p)$ is the neighborhood set of $p$, satisfying $p \notin \delta(p)$, so the CT value will be calculated according to the following equation:

$$CT(p) = \bigotimes_{p_i \in \delta(p)} \zeta(p, p_i) \quad (1)$$

Where $\otimes$ corresponds to the connected operator, and $\zeta(p, p_i)$ reflects the grayscale relations between $p$ and its neighborhood point $p_i$, $p_i \in \delta(p)$

$$\zeta(p, p_i) = \begin{cases} 1 & if I_p < I_{p_i} \\ 0 & otherwise \end{cases} \quad (2)$$

The connected manner of $\otimes$ is specified by manual. This paper connects each result from left to right, top to bottom. An example of computing CT is shown in Fig. 1:

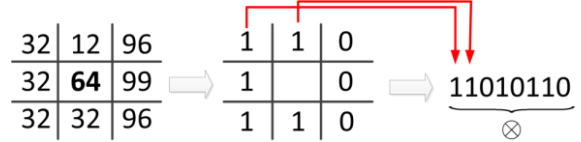

**Fig. 1** Computation Method of CT Values

From the calculation method of CT values above, we know that a CT value is composed by eight numbers from 0 to 1, which can be understood as a binary encoding pattern. For convenience, these binary CT values belonging to the same ranges can be converted to a decimal value. The calculation method is as follows:

$$CT_{10}(p) = \sum_{i=0}^{P-1} 2^i \zeta(p, p_i) \quad (3)$$

where $P$ is the number of points in $\delta(P)$. Just like Fig. 1 shows, the eight CT values can be converted to 214 based on Eq. (3).

In an image $I$, according to Eq. (3), a CT image $I'$ is calculated. And then $I'$ can be divided into M blocks (blocks may overlap or not overlap) and a local histogram in each block can be obtained ($2^P$ dimension). Meanwhile, Wu put all blocks together to form a CENTRIST feature vector. Therefore, the dimension of the CENTRIST feature vector is $2^P \times M$ [12].

## 2.2 T-CENTRIST

According to Eq. (1), binary CT pattern only considers a single pixel brightness variation between its neighboring pixels and does not take the relationship among neighboring pixels into account. Therefore, binary CT pattern will lose some image information in the conversion process. For example, in Fig. 1, according to Eq. (2), two binary CT values can be obtained ($\zeta(64,32)=1, \zeta(64,12)=1$) by comparing the center point 64 with the previous line points 32, 12 respectively. In the calculation process, we note that these two binary CT values do not reflect the brightness changes between 32 and 12. In the whole image, only one of these two points as the center point and the brightness change between the two points can be reflected. To solve this problem, this paper proposes a new Ternary Census Transform (TCT) method where Eq. (1) and (2) can be transformed as follows:

$$TCT(p) = \bigotimes_{p_i, p_j \in \delta(p)} \xi(p, p_i, p_j) \quad (4)$$

$$\xi(p, p_i, p_j) = \begin{cases} -1 & if I_p < min(I_{p_i}, I_{p_j}) \\ +1 & if I_p > max(I_{p_i}, I_{p_j}) \\ 0 & otherwise \end{cases} \quad (5)$$

where $x$ corresponds to the center point, and $p_i, p_j$ are adjacent points of $p$, satisfying $p_i, p_j \in \delta(p); i, j = 0, 1, ..., P-1$. As shown in Fig. 2, for the statistical center point 64, we compute three pixels combinations (Fig. 2-(a)) according to Eq. (5).

The first and second computations are the pixels in red and blue triangle respectively. Finally, the local ternary patterns belonging to 64 can be obtained (Fig. 2-(b)). The TCT value is $10(-1)1(-1)110$.

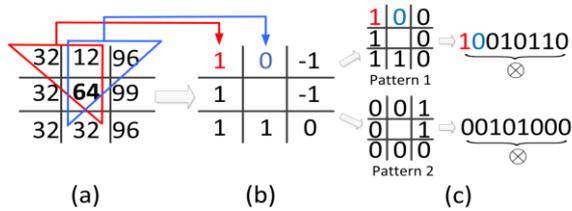

**Fig. 2** Sample of T-CENTRIST Feature Extraction

After TCT, $3^P$ patterns can be obtained to represent the human body contours. Though it can reflect brightness variation more accurately between a point and its neighboring points, the encoding pattern is not a standard binary encoding, which is not directly converted to a decimal encoding pattern in accordance with Eq. (3). In order to solve this problem, this encoding pattern for the TCT can be decomposed to two sub-patterns: Pattern 1 and Pattern 2 (Fig. 2 - (c)) and these two are calculated as follows:

$$\xi_1(p, p_i, p_j) = \begin{cases} 1 & if \xi(p, p_i, p_j) = 1 \\ 0 & otherwise \end{cases} \quad (6)$$

$$\xi_2(p, p_i, p_j) = \begin{cases} 1 & if \xi(p, p_i, p_j) = -1 \\ 0 & otherwise \end{cases} \quad (7)$$

Through Eq. (6) and (7), a TCT pattern can be converted to two standard binary patterns, which will produce a relatively sparse and large dimension histogram, increasing the computational difficulty. In order to reduce the dimensions of feature extraction, we use the idea of "uniform pattern" to reduce the types of TCT patterns[20-21]. This proposed method is called Uniform Ternary Census Transform (UTCT) whose calculation process is as follows: after the conversion and decomposition of TCT sub-patterns, if it contains at most two 0 to 1 (or 1 to 0) transitions in the corresponding binary sequence, this binary sequence will be called uniform pattern. If not, it will be called hybrid pattern.

In this paper, we apply the following formula to determine whether a binary sequence is the uniform pattern:

$$G_t(p) = \sum_{i=0}^{P-1} |\xi_t(p, p_i, p_{(i+1)\%P}) - \xi_t(p, p_{(i+1)\%P}, p_{(i+2)\%P})| \quad (8)$$

Where $t = 1, 2$ corresponds to the two different sub-patterns and % reflects modulo operation. In the case of $G_t(p) \leq 2$, the corresponding sub-pattern of $p$ is the uniform pattern. The calculation method of UTCT is as follows:

$$UTCT_t(p) = \begin{cases} \sum_{i=0}^{P-1} 2^i \xi_t(p, p_i, p_j) & if G_t(p) \leq 2 \\ P(P-1) + 2 & otherwise \end{cases} \quad (9)$$

Obviously, after UTCT conversion, the number of sub-patterns reduces significantly. Although UTCT also loses part of image information, Ojala et al. have pointed out that "uniform pattern" retains the majority of valid information in the literature, and has strong classification ability [21].

In order to calculate the T-CENTRIST feature in an image $I$, we first need to calculate two converted images $I_1, I_2$ according to Eq. (9). Second, $I_1, I_2$ can be divided into M blocks respectively, in each of which a local histogram ($P(P-1) + 2$ dimension) can be obtained. At last, all local histograms can be connected to form the T-CENTRIST feature. It should be noted that the histogram on the same block in $I_1, I_2$ will be connected as a part of the T-CENTRIST feature vector. Therefore, the dimension of the T-CENTRIST feature is $(2P(P-1) + 4) \times M$.

## 3 Fast Pedestrian Detection Framework Based on T-CENTRIST

In this paper, we use T-CENTRIST to realize a fast pedestrian detection framework. The framework of the training and testing process is shown in Fig. 3:

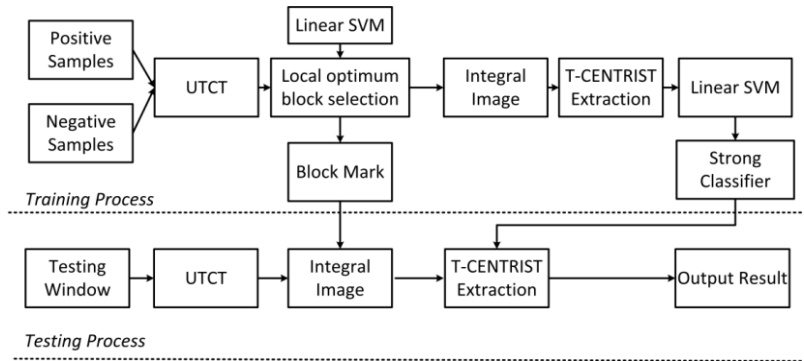

**Fig. 3** Framework of pedestrian training and testing

In the framework, two technologies are mainly applied to improve the detection accuracy. One is local extension block used to obtain better expression of body contours. The other is the integral image, which is introduced to accelerate the calculation of the T-CENTRIST feature [13].

### 3.1 Local Extension Block

In the training or testing process, the image should be divided into a series of blocks in a specified step before the T-CENTRIST feature calculation. The existing block approach is basically in accordance with the regular rectangular grid [6,7,9,13]. However, this method can not only epitomize the human's contour information, but also add much useless information, which lead to the contour features of the pedestrian being weakened, affecting the feature extraction and reducing the detection accuracy. To solve these problems, this paper applies the idea of local extension block to the segmentation of images. Five kinds of extension block structures are shown in Fig. 4.

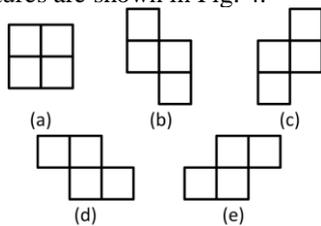

**Fig. 4** Extension block and four sub-block

When scanning the training sample set in accordance with the current specified step, we employ linear SVM to train all blocks with five different types blocks (Fig. 4) for segmentation respectively to determine the local optimal block structure. The local optimal block is the extended block with the highest detection accuracy in five different blocks and will be marked to directly extract the corresponding block in the detection process.

### 3.2 Features of the Integral Image

In pedestrian detection, using the integral image can reduce repetitive calculations, and then improve the detection speed. In [13], the author has given the calculation method and derivation process of the integral image. Base on this, a simple modification can be applied to adapt to our improvements in feature extraction and block segmentation.

According to the literature [13], we establish two auxiliary images $A_t(x, y)$. The formula $A_t(x, y)$ is:

$$A_t(x.y) = \sum_{i=1}^{n_x} \sum_{j=1}^{n_y} w_{i,j}^{UTCT_t((i-1)h_s+x,(j-1)w_s+y)} \quad (10)$$

Where $t = 1, 2$ corresponds to two sub-patterns, $h_s \times w_s$ is the size of each sub-block, $n_x \times n_y$ is the number of all blocks.

When four upper-left coordinates of a block $(t_1, l_1), (t_2, l_2), (t_3, l_3), (t_4, l_4)$ are given, we can get T-CENTRIST features by using two auxiliary images created by Eq. (10).

The corresponding block is calculated in the following equation:

$$F_t = \sum_{i=1}^{4} \sum_{x=2}^{h_s-1} \sum_{y=2}^{w_s-1} A_t(t_i + x, l_i + y) \quad (11)$$

With the integral image above, it can effectively reduce time of the feature calculation.

## 4 Experiment and Analysis

### 4.1 Experimental Preparation

To verify the effectiveness of our proposed algorithm, we test it by using INRIA pedestrian dataset [9] and some real scenes. The INRIA pedestrian dataset collected by Institut national de recherche en informatique et en automatiqueas can be used for detection of upright people in images and videos in the research work. As an important dataset, INRIA pedestrian dataset contains two parts: training set and testing set. The training set is used to train classifiers, including 614 positive images and 1218 negative images. The testing set is applied to test the performance ability, which holds 288 positive images and 453 negative images.

In our experiment, we extract some positive training samples and negative samples from the INRIA pedestrian training set. The training set includes 2000 positive samples (full-body pedestrians denoted by $P_{Train}$) and 5000 negative samples (non-pedestrians denoted by $N_{Train}$). The size of the training sample is limited to 36×72 pixels. Meanwhile, the other sample set, which contains 1000 positive samples and 1000 negative samples (full-body pedestrians and non-pedestrians denoted by $P_{Test}$, $N_{Test}$ respectively), is applied for testing our proposed method.

*4.2 A Comparison with CENTRIST*

In the literature [13], the authors have used an intuitive method to verify the classification performance of the CENTRIST, which is superior to HOG features and LBP features [9,21]. This paper will also use the similarity score mentioned in [13] to compare the classification performance of T-CENTRIST with that of CENTRIST.

For any given two samples, we can use the similarity score to estimate the similar degree of the two samples. In this paper, the calculation method of the similarity score will use Histogram Intersection Kernel (HIK) [22]. We need to calculate the histogram intersection distance of two histograms first. The computation method is:

$$D(M,N) = \frac{\sum_{i=1}^{K} \min(M(i), N(i))}{\sum_{i=1}^{K} M(i)} \quad (12)$$

where $M, N$ reflect two histograms with $K$ bins, and $M(i), N(i)$ is the $i$-th bin in $M, N$ respectively.

For any training sample, we compute the similarity score between $s$ and all other examples. Assume that $s_{in}$ is the most similar sample in the same sample set, and $s_{out}$ is the most similar sample in a different sample set. The similarity score can be calculated by:

$$Diff_s = D(s, s_{in}) - D(s, s_{out}) \quad (13)$$

In Eq. (13) above, if $Diff_s$ is positive and large, the similarity degree between $s$ and $s_{in}$ will be greater than that between $s$ and $s_{out}$.

We randomly extract 2000 positive samples and 2000 negative samples in $P_{Train}$ and $N_{Train}$ to compare the similarity scores. The results are shown in Fig. 5

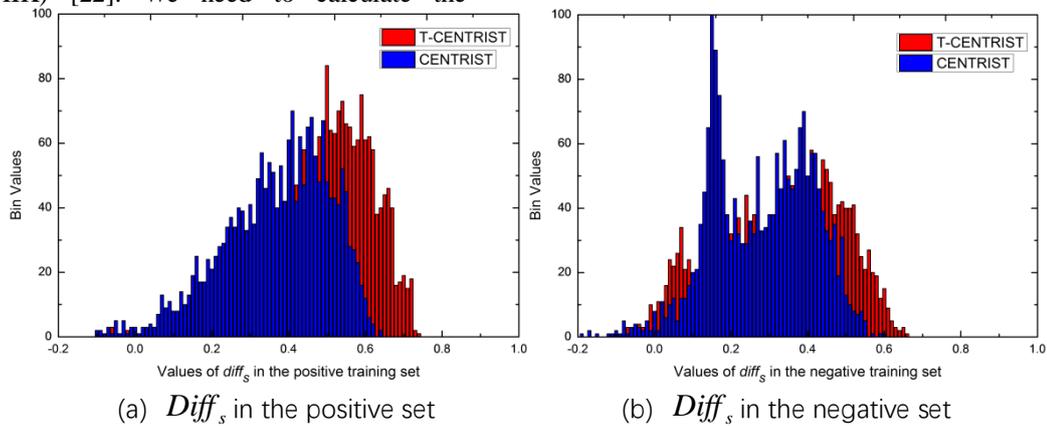

(a) $Diff_s$ in the positive set    (b) $Diff_s$ in the negative set

**Fig. 5** Similarity scores between T-CENTRIST and CENTRIST

Fig.5-(a) shows the histogram of similarity scores in the positive samples, and Fig. 5-(b) shows the same experiment output in the negative samples. In the statistical results, the percentages of $Diff_s$ in different training sets, satisfying $Diff_s < 0$, are 0.35% (positive samples) and 1.4% (negative samples) respectively by using our proposed T-CENTRIST. Under the same conditions, the percentages are 1.2% and 1.7% with CENTRIST. In addition, comparing with CENTRIST, we can obtain more values that close to 1 when using T-CENTRIST. That means that the similarity scores of T-CENTRIST are bigger than those of CENTRIST in most cases.

*4.3 Pedestrian Detection Experiment*

In the previous experiments above, we have verified that T-CENTRIST is better than CENTRIST in reflecting the characteristics of the body's silhouette. The following experiments are implemented to verify the effectiveness of our proposed pedestrian detection framework based on T-CENTRIST. A strong detector can be obtained through applying linear SVM twice on the training set. Then we compare our proposed detection framework with some existing algorithms. The results are as follows:

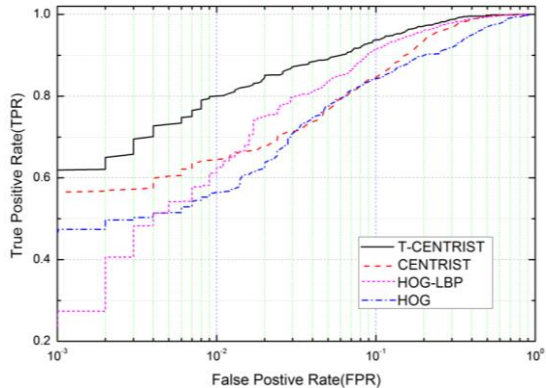

**Fig. 6** ROC for Classification

Fig. 6 shows the performance of the different classifiers. Each method for the comparison is described in [9, 12, 15]. As what can be seen from the results, our method achieves a highest detection rate of 80.2% at $10^{-2}$ FPR and 93.7% at $10^{-1}$ FPR, which is better than other state-of-the-art approaches. The detection rate in the case of using T-CENTRIST has enhanced approximately 15% than CENTRIST, higher than HOG-LBP and HOG, which means that our proposed method has effectively improved the detection rate.

## 5 Conclusion

In this paper, we first propose a new pedestrian extraction method (named T-CENTRIST) based on CENTRIST. Compared with traditional CENTRIST, T-CENTRIST not only takes the relationship between each pixel and its neighbors into account, but also considers the relationship among adjacent pixels. With such improvements, the extracted pedestrian contour information will have a stronger expression and better performance. Second, this paper presents a fast pedestrian detection framework based on T-CENTRIST. We apply local extension block and the integral image to the proposed framework. On the one hand, the framework can exclude non-human pixels as much as possible. On the other hand, it can enhance the speed for pedestrian detection by applying the integral image. At last, experimental results based on INRIA pedestrian dataset show the advantages of classification performance by using our techniques over those three other methods. Currently, our method does not make use of any other cues, such as stereo and flow. When these cues are available, they should help improve the detection performance. Therefore, the future work will focus on integrating multi-cues with our framework to improve the detection accuracy.

## *Acknowledgement*

This work is supported by PhD research startup foundation of Northeast Electric Power University under Grant no. BSJXM-2017210.